# Segment Anything Model for automated image data annotation: empirical studies using text prompts from Grounding DINO

Fuseini Mumuni[1] and Alhassan Mumuni[2]

**Abstract**

Grounding DINO and the Segment Anything Model (SAM) have achieved impressive performance in zero-shot object detection and image segmentation, respectively. Together, they have a great potential to revolutionize applications in zero-shot semantic segmentation or data annotation. Yet, in specialized domains like medical image segmentation, objects of interest (e.g., organs, tissues, and tumors) may not fall in existing class names. To address this problem, the referring expression comprehension (REC) ability of Grounding DINO is leveraged to detect arbitrary targets by their language descriptions. However, recent studies have highlighted severe limitation of the REC framework in this application setting owing to its tendency to make false positive predictions when the target is absent in the given image. And, while this bottleneck is central to the prospect of open-set semantic segmentation, it is still largely unknown how much improvement can be achieved by studying the prediction errors. To this end, we perform empirical studies on six publicly available datasets across different domains and reveal that these errors consistently follow a predictable pattern and can, thus, be mitigated by a simple strategy. Specifically, we show that false positive detections with appreciable confidence scores generally occupy large image areas and can usually be filtered by their relative sizes. More importantly, we expect these observations to inspire future research in improving REC-based detection and automated segmentation. Meanwhile, we evaluate the performance of SAM on multiple datasets from various specialized domains and report significant improvements in segmentation performance and annotation time savings over manual approaches.

## 1 Introduction

Image data annotation is one of the most important tasks in computer vision and image analysis. More generally, data annotation is a key aspect of supervised machine learning. Unfortunately, the process is one of the most laborious and time-consuming tasks in the machine learning system development pipeline. It is particularly challenging for dense annotation tasks, especially for image segmentation which requires pixel-level precision. Until recently, the process has been performed almost entirely manually. The past few years, however, have witnessed an explosion in the development of powerful generative artificial intelligence models which have demonstrated unparalleled ability to solve challenging machine learning problems without explicit human involvement. One such model, the Segment Anything Model, has undoubtedly provided new opportunities for automating tedious machine learning processes such as dataset annotation.

### 1.1 Segment Anything Model

The Segment Anything Model (SAM) [1] is a state-of the-art generative neural network pre-trained on a large volume of labeled image segmentation data. Owing to the quantity and diversity of its training data, the model is highly effective in zero-shot inference.

SAM's architecture comprises three key components: an *image encoder* that adopts the Vision Transformer (ViT) [2] to encode image features; a *prompt encoder* to extract embeddings from additional inputs like text, bounding boxes or points used to guide the model on the target regions; and a *mask decoder* that utilizes the image and prompt embeddings to generate output masks for the given input image. In the "Everything" mode, the model requires no prompt as it attempts to segment every distinct region in a given input image. In this mode, the model does not assign class names to segmented regions.

---

[1] Fuseini Mumuni University of Mines and Technology (UMaT), Ghana; fmumuni@umat.edu.gh
[2] Alhassan Mumuni Cape Coast Technical University, Cape Coast, Ghana; alhassan.mumuni@cctu.edu.gh

In this work, we perform empirical studies with pseudo-text prompts, which are not directly utilized by SAM but are, instead, used to generate bounding boxes for prompting the model. The text expressions directly serve as aliases of target category names.

**1.2** Grounding DINO

Grounding DINO [3] is a recently introduced model that has achieved landmark performance in open-set object detection. One of the best use cases of this model is in the tasks of automated zero-shot segmentation where used in conjunction with SAM. Grounding DINO extends DINO [4] with the incorporation of grounded pre-training. Grounding DINO uses separate encoders for extracting multi-scale image and text features, and a single cross-modality decoder that combines the extracted image and text modality features. In addition, a feature enhancer is employed for fusing multi-scale image and text features while a query selection module is used for selecting features which serve as inputs to the decoder. The selected input queries are then utilized by the cross-modality decoder to obtain relevant features from the image and text components. Queries obtained from the cross-modality decoder are used to predict bounding box coordinates for detected objects and extract corresponding labels.

Grounding DINO is capable of zero-shot object detection utilizing text inputs to predict targets of interest. The model can perform inference by utilizing text inputs in the form of desired object class names or referring expressions that describe object location, type, or appearance attributes like shape and color. This task is known as referring expression comprehension (REC). REC [5], [6], [7] is concerned with recognizing arbitrary targets in images by their natural language descriptions.

1.3 Image segmentation

*Semantic segmentation* is a computer vision task that assigns class-discriminative labels to each pixel in an image, whereas *instance segmentation* assigns unique labels to each instance of a class. These two common tasks, collectively known as *image segmentation,* enable machines to obtain rich, pixel-level information from images. This information provides detailed understanding of target location, size and precise geometry. Image segmentation, thus, has many practical applications.

In the healthcare domain, image segmentation enables disease regions to be delineated in medical images, thus, facilitating diagnosis and study of disease state and progression. Organ or tissue segmentation facilitates the study of anatomical structures. In robotics and autonomous driving, semantic segmentation facilitates navigation by demarcating scenes into drivable roads, lane markings and obstacles such as cars, pedestrians, trees and buildings. Plant disease segmentation enables early detection and assessment of the conditions and rate of spread of symptoms, and can provide critical information for intervention. Other essential applications in agriculture include pest segmentation which facilitates effective control; crop yield estimation; and crop quality assessment.

To achieve desired performance, state-of-the-art image segmentation models typically require large volumes of high quality training data. However, collecting and labelling large-scale image segmentation datasets is a major challenge. For this reason, methods of automating this process are of great research interest.

1.4 Image annotation for segmentation tasks

Supervised semantic segmentation requires pixel-level annotated masks for training. To create a new dataset, developers collect quality data, carefully create precise masks and corresponding labels. The common data annotation paradigms are manual, semi-automated or interactive, and fully automated methods. In manual annotation, a person (usually, with the aid of graphical tools) is required to provide all labels. The fully automated approach solely relies on machine learning models for annotation and, thus, excludes the human agent entirely from the labeling pipeline. Semi-automated or interactive methods reduce the role of the human agent to providing feedback towards correcting wrong annotations. Figure 1(a) illustrates these annotation paradigms. The manual approach is the most accessible annotation method as it only requires basic tools. However, this manual process of dataset annotation is a non-trivial task, requiring enormous amount of labor, and often at a huge financial cost. Moreover, given an image segmentation task that must be performed urgently, the time required for manual dataset annotation may prove prohibitive. Additionally, manual annotation methods may violate data privacy requirements in high-stake applications in medical image analysis, security, law enforcement and finance.

There are many open-source tools (e.g., LabelMe [8] and CVAT [9] ) that substantially reduce the labelling effort. These tools allow the user to delineate the boundaries of the regions of interest in a seamless manner. Some of these tools also incorporate semi-automated labelling features that do not require the user to manually mark every point along target boundaries. They usually integrate pre-trained deep learning models which need minimal human input to annotate images. To meet different usage needs, most semi-automated or interactive annotation systems accept different kinds of user inputs. For example, ScribblePrompt [10] produce masks from user inputs in the form of mouse clicks, scribbles, or bounding boxes. Other frameworks (e.g., AGUnet [11]) allow adaptation to the end-user's data through few-shot learning (i.e., learning with few examples), or learning from sparsely annotated data to perform dense volumetric segmentation on new data (e.g., 3D U-Net [12]).

Also, tools like MONAI Label [13], fastMONAI [14] and RIL-Contour [15] are designed to facilitate the simultaneous training of image segmentation models on custom data and the application of these trained models to image annotation on the target data. However, the performance of these methods on unseen data is poor. Also, even frameworks designed for zero-shot use cases still require some form of user input to successfully annotate user data. For instance, although CVAT [9] integrates the state-of-the-art zero-shot segmentation model, SAM, to enable fast image annotation where whole objects can be segmented with single points selected via mouse clicks, it still takes some human effort, even if minimal, to specify desired points in the targets. This makes it difficult to apply the method across large datasets. Moreover, when dealing with unknown targets, SAM sometimes struggles to separate pixels belonging to the regions of interest when only sparse points are provided.

1.5 Automated image dataset annotation

Automated data annotation or labeling seeks to leverage the power of artificial intelligence to speed up the annotation process. This also results in higher quality data, as human annotators are only employed in reviewing and refining incorrectly annotated samples rather than the tedious work of manually creating annotations from scratch, a process that can lead to fatigue induced errors. Automated data annotation may also produce more consistent results and reduce the element of bias in pixel classification.

Recent studies ([16], [17]) show that SAM outperforms interactive segmentation methods with a single or few points as prompts. Moreover, these studies ([16], [17], [18]) also reveal that prompting SAM with bounding boxes achieves far superior performance to the results obtained with point prompts. However, to provide prompts for SAM, these works extract bounding boxes or points from ground-truth masks, hence the approach cannot be applied to new datasets where no ground-truth labels exist. To enable automatic prompt generation for the Segment Anything Model, Ren et al. [19] propose the Grounded SAM framework which leverages Grounding DINO to extract bounding boxes from target objects based on text inputs.

Despite the promise of this use case, its scope is limited because in most practical applications in fields like robot vision, medical image analysis, and precision agriculture, desired targets may not readily be identified by category names. Thus, targets are usually arbitrary objects like defects in machines, disease symptoms or anatomical abnormalities, which may not fit in existing category names. Yet, these targets can be specified by their natural language descriptions. The referring expression comprehension (REC) ability of Grounding DINO leverages language understanding to support open-set detection of arbitrary targets. However, the REC framework fails to handle cases where the desired target is absent in the image. In this situation, the model produces alarming rates of false positive predictions. Consequently, the practical utility of this approach is limited.

1.6 Approach

In this work, we perform empirical studies on zero-shot semantic segmentation across diverse datasets using referring expressions in Grounding DINO to detect targets in specialized domains and then employing the detections as box prompts for SAM to generate masks. We analyze the size distribution of predicted bounding boxes and find that overwhelming majority of false positive detections can be filtered by their relative areas. Through filtering wrong predictions this way, we achieve significant improvement in segmentation quality. By selecting appropriate referring expression and setting a filter threshold, the whole pipeline from detection to segmentation can of targets can be applied across entire datasets in an autonomous fashion. The result is fully annotated images with masks and class labels. We report image annotation time reduction by as high as 94%. This process is completely autonomous, as the only inputs required from the user are text prompts and threshold to filter detected boxes. There is no pre-processing of the image data. A conceptual overview of the automated annotation framework is shown in Figure 1(b).

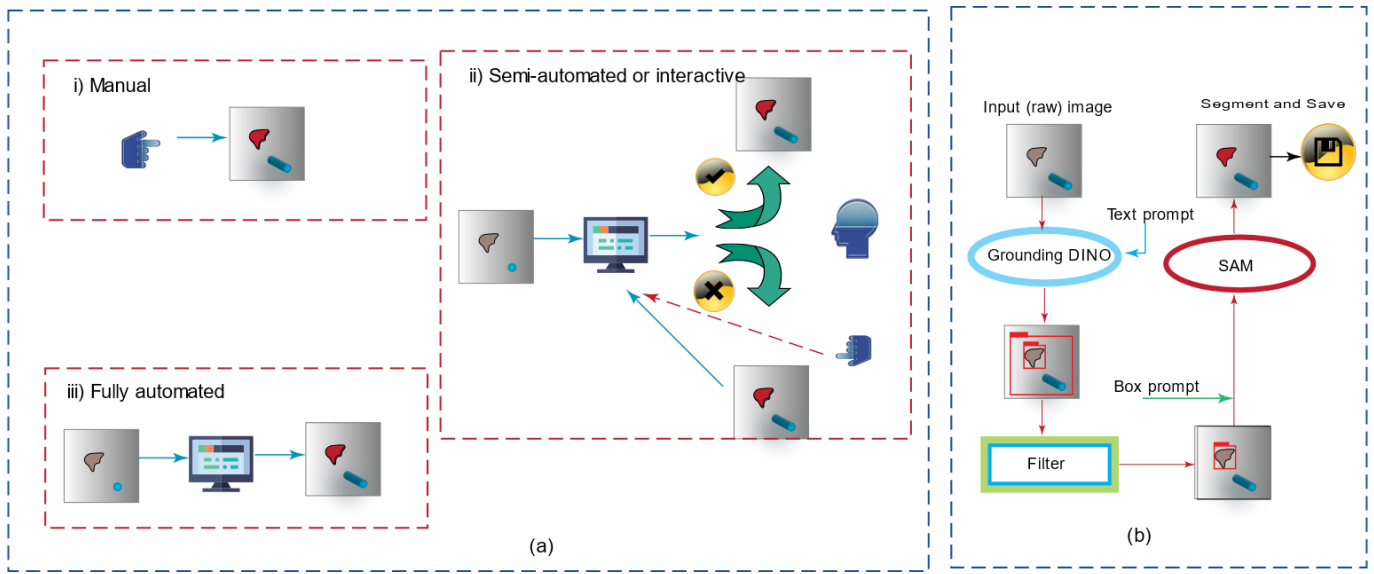

Figure 1(a) Common image data annotation paradigms; (b)A simplified representation of the Grounding DINO and SAM-based automated image annotation framework proposed in this work.

## 2. Related Work

2.1 Tackling data insufficiency with weakly-supervised semantic segmentation

One of the most widely employed solutions to the data annotation challenge is to cast the training process as a weakly-supervised semantic segmentation (WSSS) problem. In this problem setting, the data labelling effort reduces to providing weak annotations, which give approximate information on which image regions belong to the designated category or object instance. The key idea is that weak labelling is easier to perform, and a much larger volume of data can be annotated this way than with pixel-wise annotations`.

Common forms of weak annotations include bounding boxes (e.g., BoxSup [20], BBAM [21]), points (e.g., PDML [22], Point2Mask [23] ), and scribbles (e.g., Cyclemix [24]). Another poplar form of weak annotation is image-level annotation [25], [26]. Image-level annotations utilize image-level labels to indicate which object classes are present in images. Most methods leverage class activation maps (CAMs) [27] to specify approximate location information. Class activation maps highlight image regions considered to have contributed most significantly to a given class prediction. CAMs—which also include popular variants like Gradient-Class Activation Maps (Grad-CAM) [28] —generate heatmaps based on the sensitivity of the class prediction to different image regions. In typical weakly-supervised semantic segmentation, these heatmaps serve as the initial seeds for network supervision.

2.2 Segment Anything Model in WSSS

Among WSSS methods, image-level annotations are among the easiest to produce as they can readily be obtained from image classification frameworks that integrate CAM techniques. Given that these methods are simple and state-of-the-art models based on them are widely available, WSSS frameworks that utilize image-level annotations are the most popular. However, CAM generated heatmaps are only class-discriminative and do not respect individual object boundaries. Consequently, recent studies [29], [30], [31] propose to leverage SAM to enhance weakly-supervised semantic segmentation by integrating object-aware masks with class predictions. It should be noted that SAM itself is class-agnostic; it merely generates segmentation masks for objects (or their parts) without regard for their classes whereas semantic segmentation additionally requires category knowledge. Hence, CAM methods are utilized to provide class-level semantics for this group of WSSS tasks.

SAM Enhanced Pseudo Labels (SEPL) [29] leverages SAM and CAM to enhance training images for semantic segmentation. The method selects the intersection between masks predicted by SAM and CAM-generated pseudo-labels, where each segmentation mask is assigned to the CAM-generated pseudo-label with the largest area of intersection. Masks with no overlap with pseudo-labels are considered background pixels and are thus rejected. Jiang and Yang [31] propose to sample points from CAM activations which serve as point prompts for SAM whereas Yang and Gong [30] incorporates prompt learning framework that utilizes coarse-to-fine features to improve performance. The coarse features are image-level labels and CAM activations learned through image classification which then serve as seeds for learning prompts relevant to segmentation (i.e. fine features).

2.3 SAM and Grounding DINO for automated segmentation

Research interest in zero-shot image segmentation has increased tremendously owing to the immense potential of SAM in this task. In this context, several works have investigated the performance of SAM in different domains, particularly medical image segmentation [32], [33], [34], remote sensing [35], robotics [36], [37] and agriculture [38]. These works have reported impressive performance on various datasets across a wide range of image modalities. Furthermore, recent studies [17], [39] assess the utility of SAM in facilitating automated annotation of medical image datasets. They fine-tune SAM on a large collection of medical images and show that when applied in the labeling task, the method can significantly reduce the manual annotation time and improve dataset quality. However, these pipelines are not fully automated as they require additional information ground-truth data to be available. For instance, Ma et al. [39] leverage manually created sparse annotations as prompts. Huang et al. [17] rely on ground-truth annotated masks to extract points or bounding boxes for prompting SAM.

To achieve fully automated image data annotation, the entire pipeline—from prompt generation to segmentation —must progress without additional information other than specifying initial input and parameter settings to apply across an entire dataset. To enable automated prompt generation for segmentation, state-of-the-art methods incorporate auxiliary models with zero-shot generalization capability. For example, Text2Seg [40], Grounded

SAM [19] and LuGSAM [41] utilize Grounding DINO in their machine learning pipelines to provide SAM with detected boxes as prompts. In Text2Seg, J. Zhang et al. additionally assess the effectiveness of CLIP Surgery [42] in providing point prompts for automated segmentation with SAM. However, in line with other studies which report superior performance of box over point prompts, the results show higher efficacy of Grounding DINO over CLIP Surgery.

**3 Experimental Framework**

3.1 Experimental Setting

In this study, we design our experiments to investigate the characteristics of Grounding DINO detections obtained from referring expressions. We also show how to effectively leverage the referring expression comprehension of Grounding DINO and the Segment Anything Model to perform zero-shot image segmentation towards. Using seven publicly available datasets, we present empirical results in an attempt to answer the following questions:

1. Can a deeper understanding of the false positive detections of Grounding DINO help improve performance under the REC framework?
2. By how much can the segmentation performance improve by dealing with the high false positive cases of Grounding DINO detections?
3. To what extent can these improvements help to speed up image dataset annotation?

In each of the settings and for each dataset, we perform initial REC tests by conducting a few trial-and-error examples using reasonably accurate descriptions of the targets. This step is taken to determine the most appropriate text prompts that give the best detection performance. We do not perform any data pre-processing on the input images.

3.2 Datasets and pseudo-text prompts

For this work, we select a diverse set of publicly-available datasets from various domains including medical image segmentation, plant disease segmentation and seed segmentation. The datasets employed in this study are summarized in Table 1.

Here, we term the referring expressions *pseudo-text prompts* since they are not utilized directly by SAM for prediction. Instead, given an input image and a descriptive text prompt, Grounding DINO predicts bounding boxes which are used as box prompts in SAM. The choice of phrases is critical to the referring expressions comprehension task. In general, combing color and type or shape descriptors achieves good effect compared with other kinds of expressions like those relating to size. Also, referring expressions (REs) should be selected based on how well they can match relevant targets without wrongly detecting other features. A list of the expressions used for each dataset is also provided in Table 1.

In this framework, two important parameters that need to be set are the *box threshold*, which filters out bounding boxes whose confidences are lower than the set value; and the *text threshold,* which extracts only expressions with higher text similarity than the set value as labels for detections. Throughout this study, we set text threshold to 0.2 for all test cases and datasets. Box threshold is set to either 0.18 or 0.2 for all datasets except for BRACOL [43], [44] which we set to 0.16 and ISIC 2018 [45] set to 0.3. These values are informed by the detection confidences on the given datasets. As a general rule, when detection confidences for targets are high or if other objects which could match the referring expression are present in the images, it is necessary to set box threshold high enough to avoid the detection of spurious regions. On the other hand, when confidences for targets are low and there are no interfering objects, it is useful to set box threshold to a low value to reduce false negative cases. In general, too small values of box threshold can admit noisy detections while too large values can cause many valid detections to be missed.

Table 1 Summary of datasets

| Dataset | Domain | Description/Application | Plausible *RE* |
|---|---|---|---|
| **MC Chest X-Ray** [46] | Healthcare | Montgomery County (MC) Chest X-Ray dataset for lung/tuberculosis segmentation | *dark lobe* |
| **MRI Brain Tumor** [47] | Healthcare | Brain tumor detection | *white patch* |
| **BRACOL** [43], [44] | Agriculture | Brazilian Arabica Coffee leaf disease detection/ segmentation | *brown spot* |
| **ISIC 2018** [45] | Healthcare | Skin lesion segmentation dataset hosted by the international skin imaging collaboration (ISIC) | *brown patch* |
| **Plastics Segmentation** [48] | Environment | Plastic detection/ segmentation | *plastic* |
| **Tomato Seeds** [49] | Agriculture | Seeds segmentation | *brown seed* |

3.3 Evaluation Metrics

In this work, we evaluate the segmentation performance of SAM using commonly employed metrics [17], [50], [51]. For the main performance metric, we employ the Dice Similarity Coefficient (DSC or Dice). The Dice score which ranges between 0 and 1 measures the overlap between the predicted segmentation masks and the ground-truth masks, where a score of 0 indicates no overlap and 1 means complete overlap. We also include the Hausdorff Distance (HD), which measures point-wise similarity between the predicted and ground-truth masks, as well as the 95% Hausdorff Distance (HD95). These evaluation metrics enable objective comparison of segmentation performance under different settings.

**4  Results**

Although SAM can generate segmentation masks from different kinds of inputs including point and box prompts, the ability to utilize natural language descriptions as text prompts can greatly facilitate automated zero-shot semantic segmentation. The referring expression comprehension (REC) ability of Grounding DINO facilitates this use case and shows a great potential in achieving automated image data annotation at scale. However, one of the main limitations of this framework is its propensity for false positive predictions.

4.1 Characteristics of False Positive Detections

In this section we demonstrate that, despite boasting impressive performance, REC framework in Grounding DINO yields uncharacteristically poor detection results when presented with referring expressions that do not match any region in an image. Under this setting, the model frequently produces false positive detections. As shown in Figures 2 to 5 (a) and (b), arbitrary expressions that do not describe any target in an image still return high number of detections. For instance, the *referring expressions* (RE) "straight edge", "blue mask", and "blue projection", "pink stain", are all irrelevant to the images they describe, yet they return a large number of bounding boxes (false positives predictions). Even meaningless phrases like *lorem ipsum* (Figure 3(b)) produce multiple detections. This unusual behavior is mainly due to the grounding task design, where the goal is to match the referring expression to a corresponding image region, with the strict assumption that the region of interest exists. So, when the target is absent, the model returns the most logically plausible regions, which unsurprisingly, are the larger regions that contain most of the objects in the image. To obtain acceptable results, the false positive predictions must be rejected.

Fortunately, because the false positive detections follow this strikingly predictable pattern, they are easy to filter out. To illustrate this, we compute the relative areas of bounding boxes with respect to whole image areas, and compare these ratios with ground-truth bounding box areas. Figures 2 to 5 present several plots of relative areas of detected boxes against their confidence scores for different datasets. The ground-truth bounding box areas are shown as vertical (green) bars spanning all values of the confidence range (0 to 1) to signify that they do not have any specific confidence score. After analyzing the distribution of bounding box sizes for arbitrary text prompts across multiple datasets, we reveal unique patterns. Specifically, for false positive (FP) detections, the bounding boxes generally occupy large areas relative to the image size while in most cases targets of interest are limited to only small image regions. Moreover, there is usually, at most, only a marginal overlap between the regions occupied by ground-truth (green bars) and FP predictions (pink markers). For this reason, most of the erroneous detections can be filtered by their relative areas.

While previous work (e.g., [41]) attempts to filter valid detections using confidence scores, our results clearly show that these scores do not corelate positively with the fidelity of detections. That is, false positive detections often have high confidence scores while true positive detections may have low scores. To illustrate this observation, plots (a) and (b) of Figures 2 to 5 show the confidence score distributions for Grounding DINO predictions when the referred object is absent in the given image. As can be observed from plots (a) and (b) of Figures 2 through 5, the resulting false positive detections have unexpectedly high confidence scores, even higher than true positive detections shown in plots (c) in the figures. All true positive detections in (c) must be in the regions occupied by the ground-truth bounding boxes (vertical green bars). And, as shown, all such detections have generally lower confidence scores compared with (a) and (b), i.e., the FP detections obtained when the text prompt does not describe any region in the image. This observation holds true across all the datasets considered in this study. The phenomenon is easy to explain from the grounding task design: when smaller regions do not return predictions with sufficient confidence, there is increasing believe that larger enclosing bounding boxes should contain the target since it is assumed to definitely exist. In a large number of the false positive cases, the bounding box enclosing the entire image is among the predictions with the highest confidence score.

Furthermore, our results (Figures 2 to 5, plots (c)) also show that even for text queries that match targets in the given image, grounding DINO generates large number of false positive results in addition to the true positive detections, and most of these FPs also occupy larger image areas compared with the given targets. Thus, they can also be filtered.

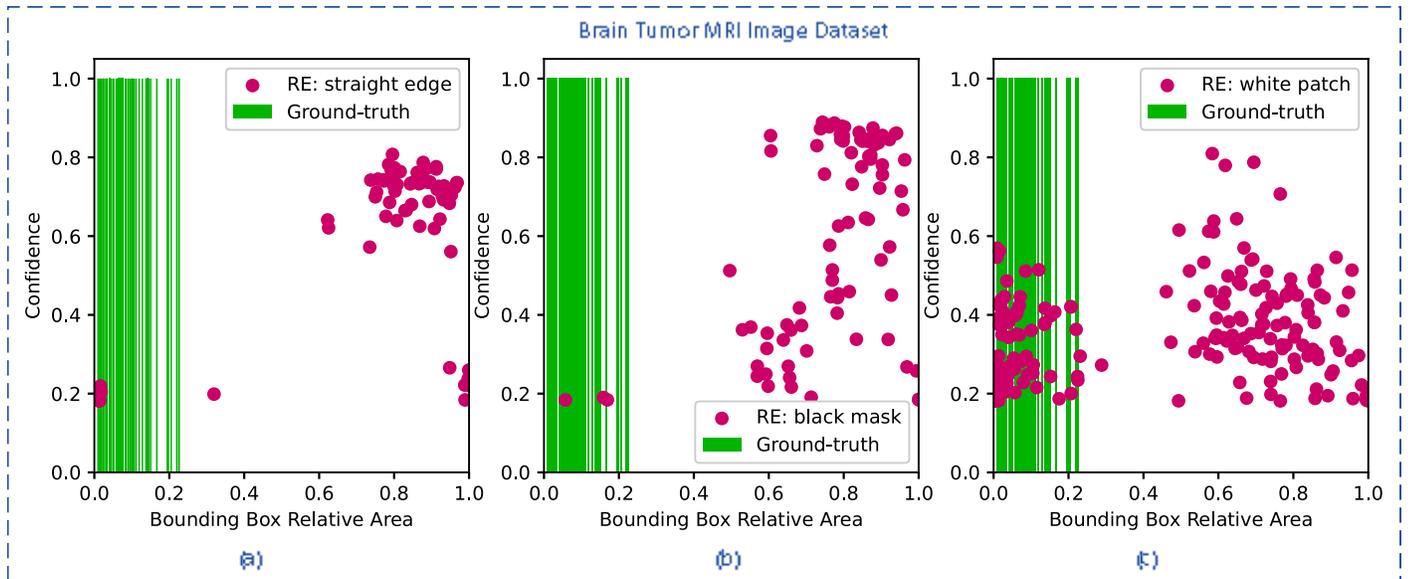

Figure 2 Relative area of detected boxes on the MRI Brain Tumor dataset [47] with different referring expressions (RE) shown against confidence scores. Ground truth bounding boxes are shown by vertical green bars. The plots demonstrate that false positive detections with higher confidence scores generally occupy large image areas, as shown in (a) and (b). However, plot (c) shows that when a plausible RE is used, targets are detected with their bounding box areas within the range occupied by ground-truths, together with FP detections with relatively larger areas.

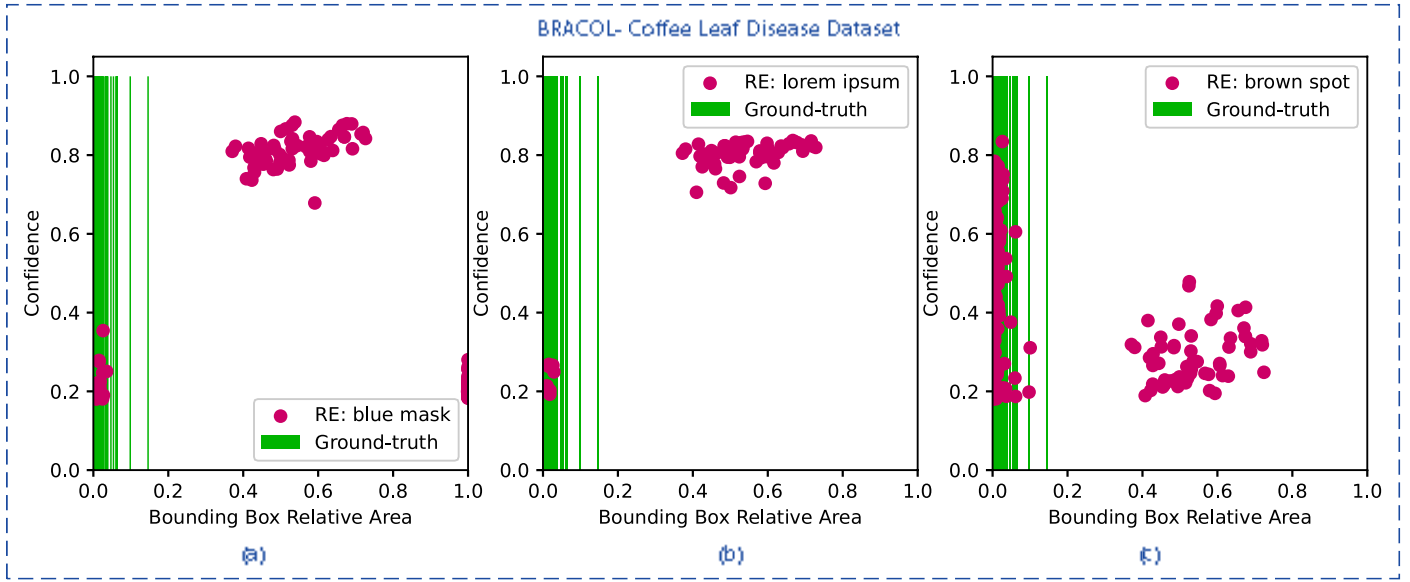

Figure 3 Relative area of detected boxes on the BRACOL dataset [43], [44] with different referring expressions (RE). False positive detections with appreciable confidence scores generally occupy large image areas, as shown in (a) and (b). In (c), a plausible RE is used, and the relative areas of detected targets within the range occupied by the ground-truths bounding boxes.

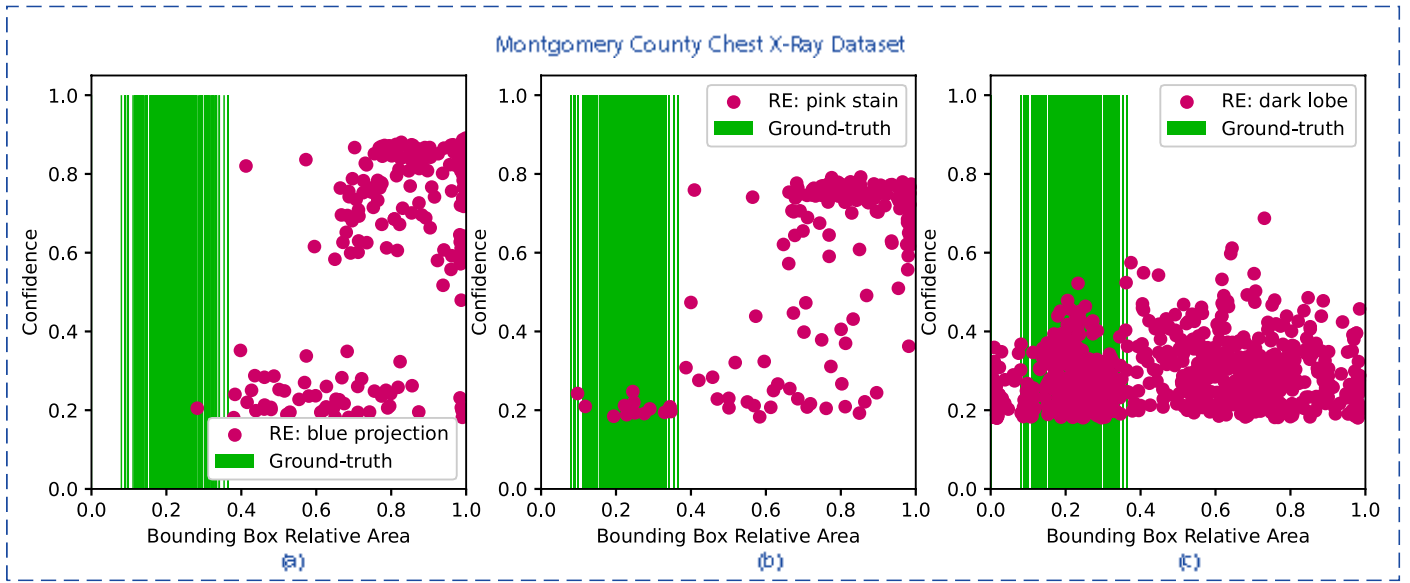

Figure 4 Relative area of detected boxes on the Montgomery County Chest X-Ray dataset [46] with different referring expressions (RE). False positive detections with appreciable confidence scores generally occupy large image areas, as shown in (a) and (b). In (c), a plausible RE is used, and the relative areas of detected targets within the range occupied by the ground-truths bounding boxes.

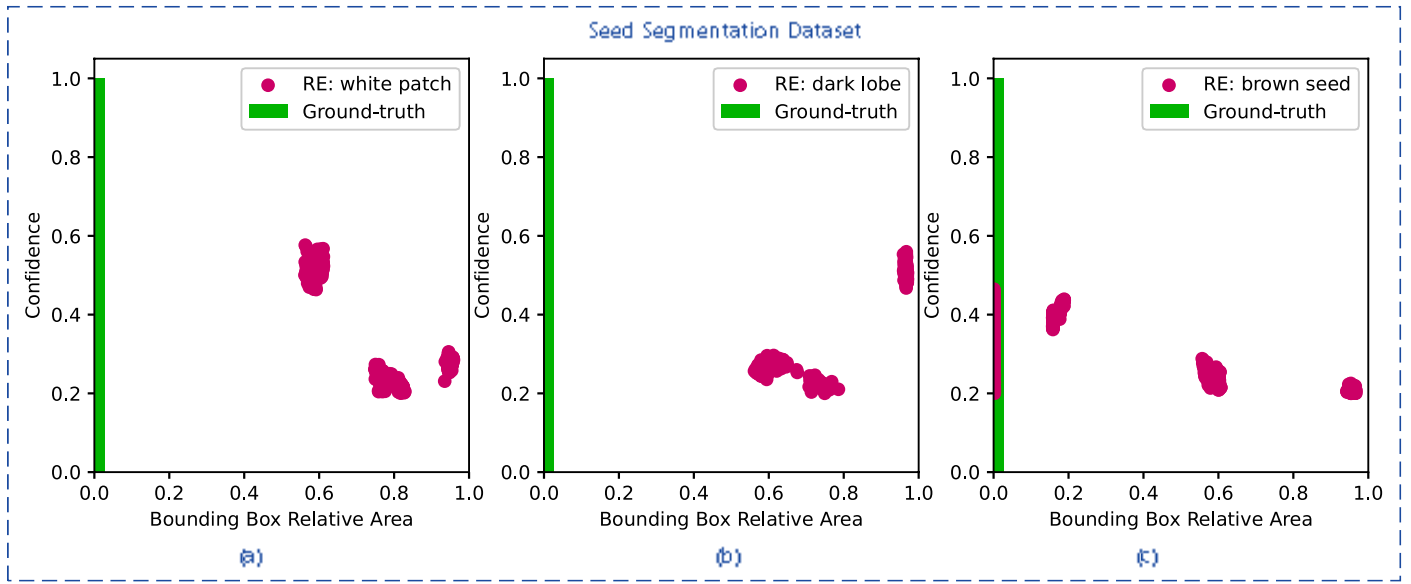

Figure 5 Relative area of detected boxes on the Tomato Seeds dataset [49] with different referring expressions (RE). False positive detections generally occupy large image areas, as shown in (a) and (b). In (c), a plausible RE is used, and the relative areas of detected targets within the range occupied by the ground-truths bounding boxes.

4.2 Determining appropriate threshold to filter detections

To successfully define size thresholds for filtering detections, it is necessary to understand broadly the features of an image that are likely to be wrongly predicted when a referring expression is supplied to Grounding DINO. Overwhelming majority of mispredictions are false positives which generally have no connection with the semantic meaning of the given expression. Even so, it is easy to anticipate the regions that would be predicted when the target is absent. This is because salient regions, particularly those spanning larger areas, are the usual candidates that are subject to false positive predictions. Therefore, it is important to select referring expressions that match only targets of interest and no other objects in the image. This way, most of the false positive detections would have their bounding boxes occupying large proportions of the image area. The detections of small features would usually have very low confidence scores and can easily be filtered by the *box threshold* parameter setting. On the other hand, when the selected expression loosely matches other image features in addition to describing the target, the detections of such features (whose sizes may not be clearly distinguishable from the targets, and so may not be easily filtered) would likely have appreciable confidence scores.

To gain more intuition on what threshold is appropriate for a given image and target, we present qualitative results of sample Grounding DINO detections based on a set of text prompts. We use both valid text prompts and arbitrarily selected expressions that that do not describe any target in the given image. We also include bizarrely meaningless phrases like *lorem ipsum* to illustrate the fact that the model almost always returns detections regardless of the referring expression supplied as input.

Figure 6 presents sample results from four of the studied datasets showing the detections produced by Grounding DINO for various referring expressions. For each of the datasets, the first two columns show the bounding boxes predicted when the referring expression is not intended to match any region in the image while the third column shows the detections generated by a plausible text prompt that matches the target. As the Figure clearly illustrates, when the target is present in the image, both true positive and false positive results may be generated. However, when the referring expression does not match any region, the predicted bounding boxes for the false positive detections generally enclose large image areas and tend to have high confidence scores. Moreover, these bonding boxes do not merely occupy random positions. In general, they tend to delineate regions of visual saliency and are objectness-aware. Owing to this behavior, it is possible to tell by observation (in most cases), which regions are likely to be predicted as false positives. Hence, selecting appropriate size threshold (i.e., largest admissible bounding box) to filter out false positive detections considers these factors in conjunction with target sizes. This choice is easy to make by inspecting sample images with larger targets and estimating the approximate ratio of the region of interest to whole image area.

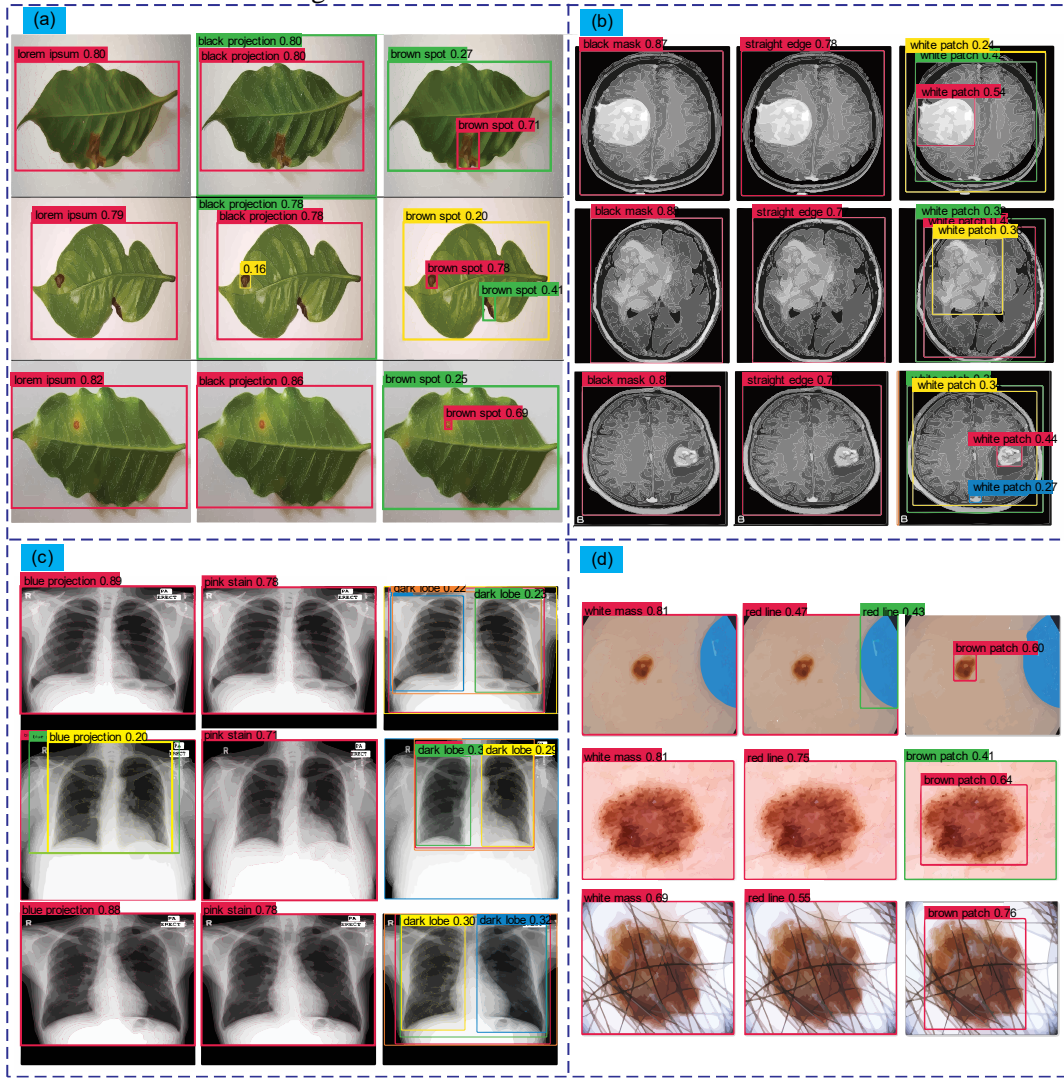

Figure 6 Sample Grounding DINO detections on four selected datasets: (a) BRACOL [43], [44]; (b) MRI Brain Tumor [47]; (c) Montgomery County Chest X-Ray [46] and (d) ISIC 2018 [45] datasets. For the first two columns of the results shown for each dataset, the target is absent in the image and all detections with appreciable confidence scores occupy large image areas. In the third columns on each dataset, plausible referring expressions are used and the results show correct detections, sometimes in addition to false positive predictions with large bounding boxes.

4.3 Zero-shot segmentation performance

Our empirical studies in Sections 4.1 and 4.2 reveal that for most of the false positive detections, the predicted bounding boxes generally occupy large image regions. Yet, in many practical image segmentation applications, the regions of interest (ROIs) are restricted to only small areas of the given image. For instance, in a typical medical image segmentation task, the target may be an organ, a tissue or even a defect in these structures, whereas the dataset may capture the entire body part that contains the target. Similarly, in plant disease segmentation datasets, the affected regions are usually captured along with whole plants or plant parts. Consequently, in all such scenarios, expected ROIs would occupy a relatively small area of the image. Therefore, using this strategy, we find that false positive detections can easily be filtered out by simply rejecting bounding boxes whose relative sizes are above a pre-defined threshold.

Based on this observation, we evaluate zero-shot segmentation performance on five publicly available datasets. We present two set of results on 1) segmentation utilizing raw Grounding DINO detections; and 2) segmentation utilizing filtered detections that exclude larger bounding boxes. The results are presented side-by-side with the selected relative area threshold (third column) applied to filter the detections.

A quantitative comparison of the raw and filtered results (Table 2) demonstrates a wide gap between the two sets of results for all the studied datasets. The BRACOL and Seed segmentation datasets record the most significant performance differences. For instance, we record improvements in DSC scores of 900% or higher on the BRACOL (*phoma* class– 936%, *miner* class –1017%) and Seed (990%) segmentation datasets. These huge differences can be explained by the relatively smaller sizes of the targets compared to the total image area. Since false positive detections occupy relatively large proportions of image area, they significantly degrade the Dice score for smaller targets. Therefore, as expected, the Montgomery County chest-x-ray and the ISIC skin lesion datasets show a comparatively lower performance gain (29.50% and 32.31%, respectively) over the unfiltered results.

Table 2 also presents selected results which exclude segmentations with very low (in this case 0.0) dice scores. The intuition is that in a typical automated data annotation task, only correctly annotated images and those with minor deficiencies would be considered. Definitely, the assumption is that a large proportion of segmentation results are of acceptable quality. Interestingly, as expected, the poor results are identified as outliers (red squares) in the Box plots of Figures 7 and 8. Most of these very poor results come from false negative detections caused by our setting of *box threshold* parameter (we set it at 0.18 or 0.2 for all the datasets) of Grounding DINO. This setting essentially filters out detections with confidence scores less than 0.18 (0.2). In cases where all valid detections are less than this threshold, the resultant Dice score is 0.

Table 2 Average values of Dice, HD snd HD95 scores across selected datasets. Dice scores range from 0 to 1 and higher values show better performance. For HD and HD95, higher values are better.

| Dataset | Category/target | Rel. Area | Metric | All results | | Selected Filtered results |
|---|---|---|---|---|---|---|
| | | | | Raw | Filtered | |
| Montgomery Chest X-Ray | lung | 0.45 | Dice | 0.5447 | 0.7054 | 0.8038 |
| | | | HD | 102.46 | 79.79 | 54.60 |
| | | | HD95 | 85.73 | 65.84 | 40.79 |
| ISIC | Skin lesion | 0.90 | Dice | 0.58 | 0.7810 | 0.8377 |
| | | | HD | 98.47 | 67.15 | 50.21 |
| | | | HD95 | 81.11 | 59.84 | 41.33 |
| Bracol | cercospora | 0.12 | Dice | 0.0933 | 0.8138 | √ |
| | | | HD | 213.66 | 42.88 | √ |
| | | | HD95 | 195.92 | 31.30 | √ |
| Bracol | miner | 0.12 | Dice | 0.0838 | 0.9362 | √ |
| | | | HD | 212.88 | 50.2932 | √ |
| | | | HD95 | 189.99 | 36.5485 | √ |
| Bracol | phoma | 0.12 | Dice | 0.0913 | 0.9459 | √ |
| | | | HD | 216.34 | 51.81 | √ |
| | | | HD95 | 195.36 | 36.02 | √ |
| Bracol | rust | 0.12 | Dice | 0.0979 | 0.6818 | 0.7171 |
| | | | HD | 168.46 | 76.76 | 70.16 |
| | | | HD95 | 144.07 | 51.97 | 45.745 |
| Plastic | plastic | 0.35 | Dice | 0.6230 | 0.7352 | 0.8254 |
| | | | HD | 220.69 | 200.30 | 141.15 |
| | | | HD95 | 177.13 | 171.92 | 112.58 |
| Seed segmentation | seed | 0.02 | Dice | 0.0994 | 0.8317 | √ |
| | | | HD | 201.39 | 58.73 | √ |
| | | | HD95 | 188.95 | 40.36 | √ |

Figures 7 and 8 summarize the experimental results in box and whisker plots. The plots indicate that filtered results achieve comparatively better performance, as shown by their consistently higher median scores. Segmentation from raw detections show a generally narrower quartile range owing to the absence of high scores. On the other hand, except for a few outliers shown by the red squares, all of the filtered detections yield segmentation results whose Dice (DSC) scores range from at least 0.4 upwards.

Also, the performance gap across different datasets is again noticeable from Figures 7 and 8. Specifically, the Dice scores computed on the various categories of the BRACOL dataset, as presented on Figure 8, show a much higher improvement in segmentation quality as a result of pre-filtering larger bounding boxes. On the other hand, Figures 7 shows the plots for four other datasets with impressive but relatively lower performance improvement resulting from filtering erroneous detections. The difference in performance improvement shown by the two plots comes from the fact that in Figure 8, the ratio of the image area occupied by the targets is relatively small. Hence, the presence of false positive detections (in the unfiltered case), which usually occupy large image areas, severely affect the segmentation results. The peculiarities can also be seen from Figure 9.

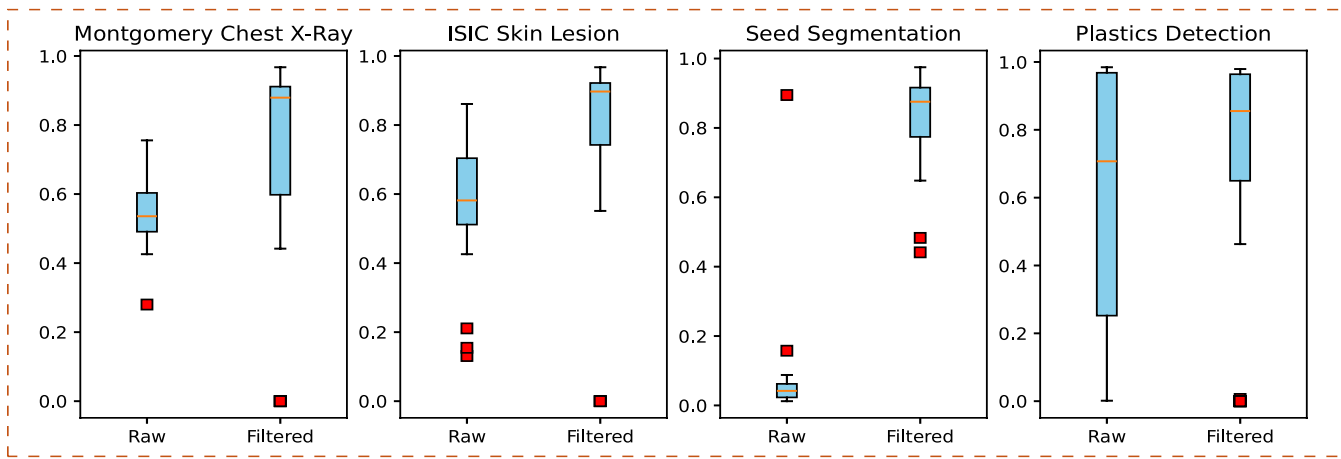

Figure 7 Box plots showing the raw and filtered detections on four different datasets

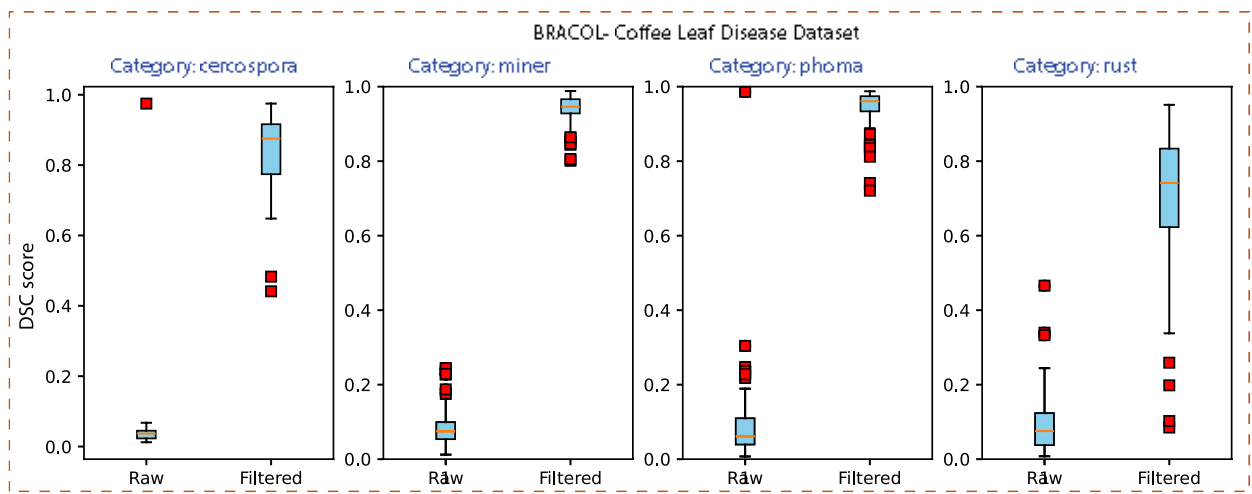

Figure 8 Box plots showing the raw and filtered detections of various categories of the BRACOL dataset

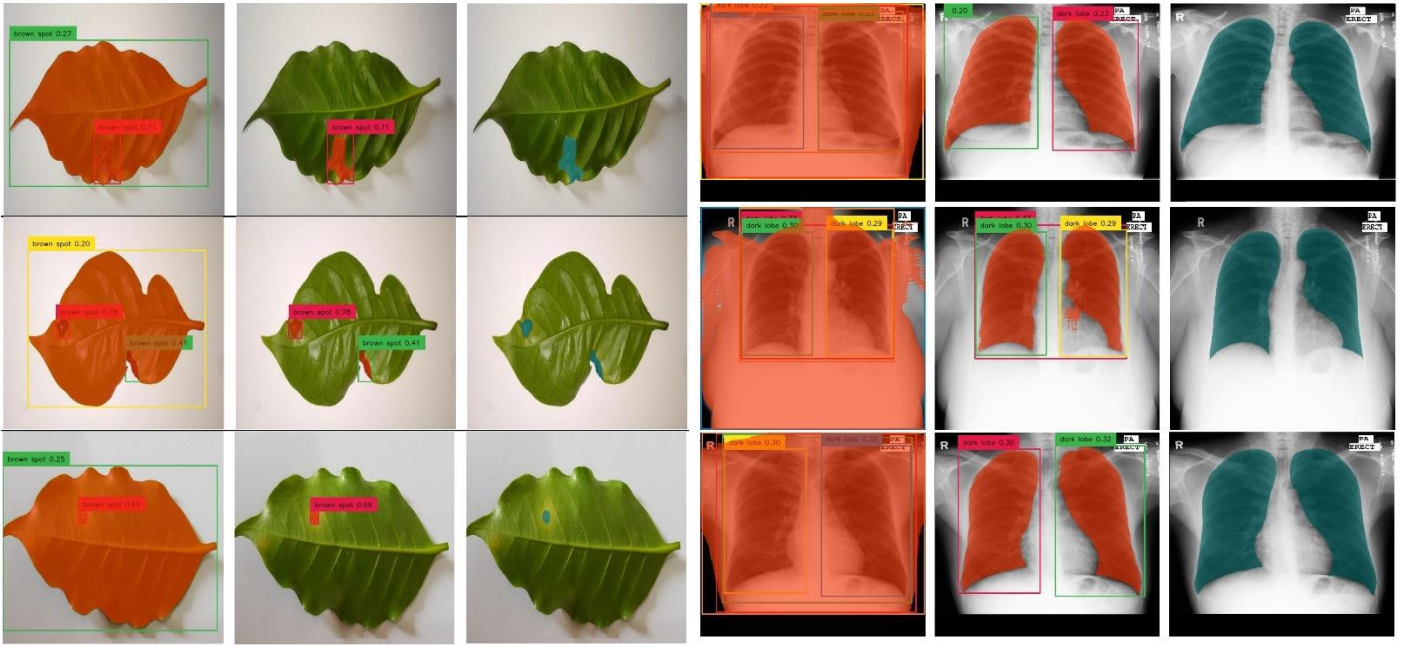

Figure 9 Qualitative assessment of the impact of filtering Grounding DINO detections on quality of SAM segmentation. The left half of the image show output for BRACOL while the right half is for the Montgomery County Chest X-Ray dataset. In each case, the first (leftmost) column represents segmentations from raw detections, middle column shows segmentation from filtered detections while the right column shows ground-truth results.

4.4 **Automated Image Annotation Performance**

In this section, we present results on the effectiveness of leveraging the capabilities of Grounding DINO and SAM to facilitate automated image annotation for semantic segmentation tasks. The task of automated image annotation is accomplished by simply providing appropriate referring expression to Grounding DINO to detect desired targets and then supply the predicted bounding boxes to SAM for segmentation. The results are subsequently verified and refined as needed. The main goal of this activity is to reduce the time needed to produce labeled semantic segmentation datasets from scratch.

To estimate the annotation time savings of automated segmentation over manual annotation, we arbitrarily select a subset of images from the datasets used in this study to form our evaluation subset. The selected subset comprises 60 images (specifically, the first 60 images when sorted alphabetically in ascending order by name) from the Montgomery County (MC) Chest X-Ray dataset and 60 images belonging to the *phoma* category of the BRACOL training set. We engage 60 human annotators (final year undergraduate students in Computer and Electrical Engineering taking Digital Image Processing course) who we divide into two groups to work on the two datasets, with each group further divided into two subgroups. The first subgroups perform manual annotation from scratch using the Computer Vision Annotation Tool (CVAT) as an aid. The second subgroups inspect and refine SAM-generated masks with the help of CVAT. Each human annotator is given a few samples of the images and corresponding ground-truths for reference. The procedure involves labeling and timing the duration on a per-image basis.

In addition to the annotation time cost, we also assess the reduction in labor achievable with SAM. Specifically, in line with common practice (see [17], [52]), we employ the Human Correction Efforts (HCE) criterion to estimate the annotation cost in terms of human effort required to annotate the given images from scratch or to refine the generated masks to reach the quality of ground-truths. While standard evaluation metrics attempt to characterize the semantic gap between ground-truth and predicted masks, Qin et al. [52] propose the HCE metric as a measure that reflects the actual human effort (for example, in terms of the number of mouse clicks) required to refine the masks to match ground-truth samples. In the context of the current work, HCE is basically the number of clicks needed to select the points that delineate target boundaries. We also include clicks used to re-position points which were initially misplaced, as well as selecting the "Polygon" tool and the mask label in CVAT interface. Apart from clicking to place points on the canvas, the annotation tool also allows users to simply move their mouse cursors continuously along the target boundary to select fine-grained points. Our annotators extensively utilize this feature, as it is faster. In this mode we approximate HCE to 1 click every second of cursor movement. In the cases where extraneous regions have been included in SAM-predicted masks, HCE considers the number of mouse clicks used to remove these redundant regions in edit mode.

The results presented in Table 3 are averaged per image for all annotators. They show a significant reduction in time and labor. In particular, automated annotation is most effective on the samples from the BRACOL dataset where it achieves an impressive time saving of about 94% and HCE improvement of about 89%. The BRACOL images are easy to segment since they have well-defined boundaries and sufficiently high contrast. In fact, most of the generated masks bare need any correction at all. In contrast, the Montgomery County Chest X-Ray dataset records a modest reduction of 33% in annotation time and 34% HCE. These contrasting results are obvious from the characteristics of the datasets. Most of the segmentations of the BRACOL images hardly need any refinement at all.

Table 3 Comparison of automated and manual annotation cost

| Dataset | Data Preprocessing? | From scratch | | With SAM | | Improvement, % | |
|---|---|---|---|---|---|---|---|
| | | HCE | Time (min) | HCE | Time (min) | HCE | Time |
| MC Chest X-Ray | No | 453 | 2.4 | 298 | 1.6 | 34 | 33 |
| BRACOL | No | 268 | 1.1 | 16 | 0.07 | 89 | 94 |

It is evident from the results presented in this section that automated image dataset annotation using SAM and REC-framework exhibits varying levels of strengths depending on the target dataset. The performance depends both on the ability to correctly detect targets and the quality of the segmentation masks that can be produced from the resulting bounding box prompts. In general, targets with distinct boundaries and high contrast found in images with minimal background clutter are both easy to detect and segment. Consequently, datasets (such as the BRACOL) which exhibit these characteristics are most suitable for automated annotation. In these cases, annotated images require negligible human effort to correct erroneous segmentation results.

## 5 Discussion

Despite the immense promise of zero-shot image segmentation utilizing Grounding DINO and the segment anything framework, the performance is still notoriously poor when applied in the automated regime to specialized domains where targets must be identified by their semantic descriptions (also known as *referring expressions*). This challenge is largely due to false positive predictions. To this end, we conduct comprehensive experiments on multiple datasets to understand the problem. Primarily, through this study we show that most of the detection errors encountered in using this framework are false positives. That is, false negative cases are generally rare. Moreover, these false positive detections do not occupy arbitrary regions, but delineate salient regions with higher affinity shown towards artifacts occupying larger areas. Consequently, we show that significant improvement in segmentation performance can be achieved by leveraging this predictable behavior to pre-filter detections by their bounding box sizes. More importantly, we expect the findings in this study to lead to further understanding and additional insights that can be leveraged to improve automated segmentation or data annotation.

A major role played by human annotators in semi-automated or interactive labeling frameworks is in resolving semantic ambiguities in images. However, with the incorporation of referring expression comprehension much of this need can be met by simply selecting appropriate natural language descriptions that match targets in the given situation. Therefore, the proposed pipeline reduces the manual effort required in data annotation and can potentially obviate the need for human involvement completely. This development can greatly enhance the overall performance of machine learning models in practical applications like agriculture, healthcare, robotics and manufacturing domains which deal with complex images or scenes with wide variability. This is because standard datasets used to train machine learning models are limited in variability and representativeness of real-world environments. Hence, these models experience performance drop when deployed in practical applications. However, as shown by the results in this study, the ability of Grounding DINO to leverage natural language descriptions greatly facilitates open-set semantic segmentation by SAM, where arbitrary targets can be identified and segmented with the aid of natural language descriptions. The ease with which this annotation process can be carried out implies that new dataset can effortlessly be created for any real-world application scenario as needed rather than relying on existing datasets which may not align well with the given domain.

Despite its promise, there are still a few limitations with the framework discussed in this work. The first and most obvious one is that the method will struggle when applied in situations where the targets are not so distinct from other features in the images. Also, in the current circumstance, human experts are still required to review every annotation produced by the automated framework to establish the correctness of the generated masks. This process adds a significant labor cost. If the model could simultaneously annotate images and estimate which predictions are likely to be unreliable, the entire pipeline could be automated. In this situation, the model can automatically accept sufficiently accurate annotations while rejecting results with high prediction uncertainty, all without the need for human involvement.

As the results presented in this work show, detection confidence scores are not reliable at all. Hence, they cannot be used as an accurate measure of prediction uncertainty. Furthermore, even if detection confidences are known, the confidence of the segmentation task must also be computed to obtain overall uncertainty estimation. Fortunately, research in uncertainty estimation in object detection and segmentation is already yielding promising results [53], [54]. One of the simplest ways that have been suggested to accomplish this is by taking the output of the final activation layer of a network as a measure of prediction uncertainty. With this approach, outputs close to 1 are considered to characterize high level of certainty that the given pixels belong to the predicted targets. A major limitation of this mechanism is that it does not account for the sources of uncertainty. With this capability, it would be possible to give higher priority to more relevant sources and disregard those that do not apply in the given situation. Despite current challenges, uncertainty estimation is advancing positively and this progress is expected to significantly increase the level of autonomy of automated image annotation pipelines.

## 6 Conclusion

Inspired by the recent progress in zero-shot semantic segmentation facilitated by the powerful Segment Anything Model, we conduct thorough empirical studies with the aim of enhancing automated segmentation tasks like data annotation. Our work focuses first on addressing one of the most important limitations of language-driven open-set object detection with Grounding DINO – false positive predictions. After filtering out these wrong detections, we provide the remaining bounding boxes as prompts to SAM. The resulting masks generated show remarkable quality. Moreover, our results demonstrate that this framework is highly effective in reducing data annotation labor and time cost.